\documentclass[conference]{IEEEtran}
\pdfoutput=1
\usepackage{cite}
\usepackage{graphicx}
\usepackage{amsmath}
\usepackage{amssymb}
\usepackage[letterpaper, left=1.01in, right=1.01in, bottom=1in, top=0.75in]{geometry}

\newcommand{\RN}[1]{%
  \textup{\uppercase\expandafter{\romannumeral#1}}%
}

\IEEEoverridecommandlockouts \IEEEpubid{\makebox[\columnwidth]{ 978-1-5386-3531-5/17/\$31.00~\copyright~2017 IEEE \hfill} \hspace{\columnsep}\makebox[\columnwidth]{ }}
\begin{document}
\title{Deep Learning for Malicious Flow Detection}

\author{\IEEEauthorblockN{Yun-Chun Chen\IEEEauthorrefmark{1},
Yu-Jhe Li\IEEEauthorrefmark{2}, Aragorn Tseng\IEEEauthorrefmark{2}, and
Tsungnan Lin\IEEEauthorrefmark{1}\IEEEauthorrefmark{2}\IEEEauthorrefmark{3}}

\IEEEauthorblockA{\IEEEauthorrefmark{1} Department of Electrical Engineering, National Taiwan University}
\IEEEauthorblockA{\IEEEauthorrefmark{2} Graduate Institute of Communication Engineering, National Taiwan University}
\IEEEauthorblockA{\IEEEauthorrefmark{3} Cybersecurity Technology Institute, Institute for Information Industry}
}

\maketitle

\begin{abstract}
Cyber security has grown up to be a hot issue in recent years. How to identify potential malware becomes a challenging task. To tackle this challenge, we adopt deep learning approaches and perform flow detection on real data. However, real data often encounters an issue of imbalanced data distribution which will lead to a gradient dilution issue. When training a neural network, this problem will not only result in a bias toward the majority class but show the inability to learn from the minority classes.

In this paper, we propose an end-to-end trainable \textit{Tree-Shaped Deep Neural Network} (TSDNN) which classifies the data in a layer-wise manner. To better learn from the minority classes, we propose a \textit{Quantity Dependent Backpropagation} (QDBP) algorithm which incorporates the knowledge of the disparity between classes. We evaluate our method on an imbalanced data set. Experimental result demonstrates that our approach outperforms the state-of-the-art methods and justifies that the proposed method is able to overcome the difficulty of imbalanced learning. We also conduct a partial flow experiment which shows the feasibility of real-time detection and a zero-shot learning experiment which justifies the generalization capability of deep learning in cyber security.
\end{abstract}

\IEEEpeerreviewmaketitle

\vspace {-1.5mm}

\section{Introduction}

Cyber security has become an important issue that cannot be overlooked. Recently, a notorious type of Ransomware, WannaCry, has posed a threat to the entire cyber society. This Ransomware has appeared numerous kinds of variants so far and the third version whose targeted vulnerability differs from others has come into existence as well.

Traditional methods to defend these threats are signature oriented, which means that anti-virus software will not be able to detect them if malware changes its behavior. On the other hand, real-time detection is also a concern when applying anti-virus software.

Recently, many literature \cite{alqurashi2016comparison}, \cite{demme2013feasibility} proposed different methods to detect malware. Even though the methods mentioned in these literature can achieve a high accuracy, we discover that the precision, which is also an important index in machine learning tasks, of the proposed methods aren't high. To seek for a substantial improvement on the precision of malware detection and perform a detailed classification, we construct a neural network-based model based on our understanding regarding neural network's potential and generalization capability.

Deep learning has an excellent performance in many aspects such as computer vision task \cite{NIPS2012_4824} and speech recognition task \cite{mikolov2010recurrent} to name a few. With the emergence of various types of neural network architectures and different training mechanisms, deep learning has become a feasible and powerful approach when handling problems with complexity.

Data is an important issue in building a robust model. Unlike computer vision tasks that data are available on the Internet with ease since there are numerous open data sets, on the contrary, to the authors' best knowledge, the task we are conducting does not have any malware open data set and therefore the malware data samples used in this task are collected by authors. Under this situation, the collected data will often result in an imbalanced data distribution.

Imbalanced data distribution occurs quite frequently in many machine learning related tasks \cite{comcsa2014adaptive}, \cite{comcsa2014scheduling} and this issue will become a major challenge in training neural networks due to the gradient dilution issue. Besides, the number of malware we obtained is even less than we expected. How to construct a robust and unbiased model becomes a challenging problem in our task.

To tackle the above-mentioned challenge, we propose an end-to-end trainable \textit{Tree-Shaped Deep Neural Network} (TSDNN) along with a \textit{Quantity Dependent Backpropagation} (QDBP) algorithm which incorporates the knowledge of the disparity between classes and classifies the data layer by layer. We evaluate the effectiveness of our method on a considerably imbalanced data set. The main contributions of this paper are summarized as the following:

\begin{itemize}
\item We propose an end-to-end trainable TSDNN model which classifies the data in a layer-wise manner. It is shown experimentally to perform far beyond the multi-layer perceptron architecture on an imbalanced data set.
\item We propose a QDBP algorithm which incorporates the knowledge of the disparity between classes and overcome the difficulty of imbalanced learning. Experimental result shows that our method outperforms the state-of-the-art approaches on an imbalanced data set whether in accuracy or precision.
\item We further show the feasibility of real-time detection by executing a partial flow experiment and we also conduct a zero-shot learning experiment which demonstrates the generalization performance of deep learning on cyber security.
\end{itemize}

The rest of the paper is organized as follows. The next section presents the state-of-the-art methods to tackle imbalanced data issue. Section \RN{3} introduces the terminologies used in this paper and describes the phenomenon of gradient dilution. The proposed algorithm and the architecture of neural network are elaborated in section \RN{4}. Section \RN{5} presents the experimental framework and provides comprehensive experimental results. Eventually, we conclude this paper in section \RN{6}.

\section{Related Work}
Imbalanced data distribution is an inevitable issue when performing classification tasks. There are many literature \cite{comcsa2014adaptive}, \cite{comcsa2014scheduling} aimed at tackling the challenge, and the proposed methods can be categorized into two main strategies, data manipulating techniques and training mechanism adjustment.

\subsection{Data Manipulating Technique}
There are two types of data manipulating techniques, the oversampling method \cite{pears2014synthetic} and the undersampling method \cite{liu2009exploratory}. Oversampling is a method that constantly re-samples the minority class into the training data set. Undersampling, on the other hand, is a method that randomly eliminates the data of the majority class. Even though these two methods are opposed to each other, both of them are aimed to adjust the data distribution and mitigate the imbalanced data issue.

\subsection{Training Mechanism Adjustment}
Incremental learning is a feasible method to mitigate imbalanced data issue \cite{kulkarni2014incremental}. The objective of incremental learning is intended to learning from the newly introduced data without forgetting past memories. The model usually trains on a small data set initially and increases the size of training data set gradually.

\section{Gradient Dilution}
Training neural networks is a gradient-based method which updates the parameters according to the partial derivatives computed with respect to the cost function. When computing gradient in each epoch, the number of the gradient contributed by each class equals to the vector summation over the training data of each class. If the number of the data of the majority class is much greater than that of the minority class, the model will tend to update the parameter toward the majority class. To understand the phenomenon, we have to analyze the computation of gradient when training a neural network.

For each epoch, every training sample will be trained by the model once. Namely, the model will update its parameters with respect to each data once in every epoch. The total gradient of each epoch will be the sum of all gradients contributed by each individual training data and the mathematical formula is given by:

\vspace {-4mm}

\begin{equation}
    \frac{\partial Loss}{\partial\theta} = \sum_{i=1}^{N} \frac{\partial Loss_i}{\partial\theta}
\end{equation}

\vspace {-1.5mm}

where $\frac{\partial Loss}{\partial\theta}$ refers to the total gradient of each epoch with respect to $\theta$, $\frac{\partial Loss_i}{\partial\theta}$ refers to the gradient contributed by the $i^{th}$ training data with respect to $\theta$, and \(N\) represents the number of training data samples in each epoch.

If we reorder equation (1), and group the gradient contributed by the same class together, we will have

\begin{equation}
    \frac{\partial Loss}{\partial\theta} = \sum_{M_i \in M}^{} \sum_{A_{ik} \in M_i}^{} \frac{\partial Loss_{A_{ik}}}{\partial\theta}
\end{equation}

where \(M\) represents the training data set, \(M_i\) refers to the $i^{th}$ class of \(M\), \(A_{ik}\) is the $k^{th}$ data of \(M_i\), and \(Loss_{A_{ik}}\) is the loss contributed by data \(A_{ik}\).

From equation (2), we can observe that the number of updates of parameter $\theta$ with respect to each class depends on the size of each class. Once the number of the data disparate between classes severely, the model will tend to bias toward the majority class since the total gradient will be dominated by the gradients contributed by the majority class in terms of frequency. This phenomenon will result in the insensitivity of the model toward the minority classes because the model seldom updates the parameter with respect to the minority classes. Metaphorically, the gradients contributed by the minority classes fade as if it were diluted by that contributed by the majority class. This phenomenon is called the gradient dilution.

For instance, considering a binary classification on an imbalanced data set where positive class contains 10,000 pieces of data whereas only 10 pieces of data belongs to the negative class. For every epoch, the model will update the parameters toward the direction of the positive class 10,000 times while only updates 10 times toward the direction of the negative class. Under this scenario, the model will tend to bias toward the positive class since the model updates more frequently toward the direction of the positive class.

The effect of the gradient dilution depends on the number ratio of the data between classes. If the ratio is large to some extent, $\frac{216,015}{45}$ in our task, the model will be more likely to update toward the majority class and result in an inability of the model to learn from the minority classes. To mitigate the impact of the gradient dilution without adjusting the distribution of the data, we propose a QDBP algorithm which takes the disparity between classes into consideration, adjust the sensitivity of the model with respect to each class, and is able to overcome this difficulty.

\begin{figure*}[!t]
    \centering
    \includegraphics[scale=.47]{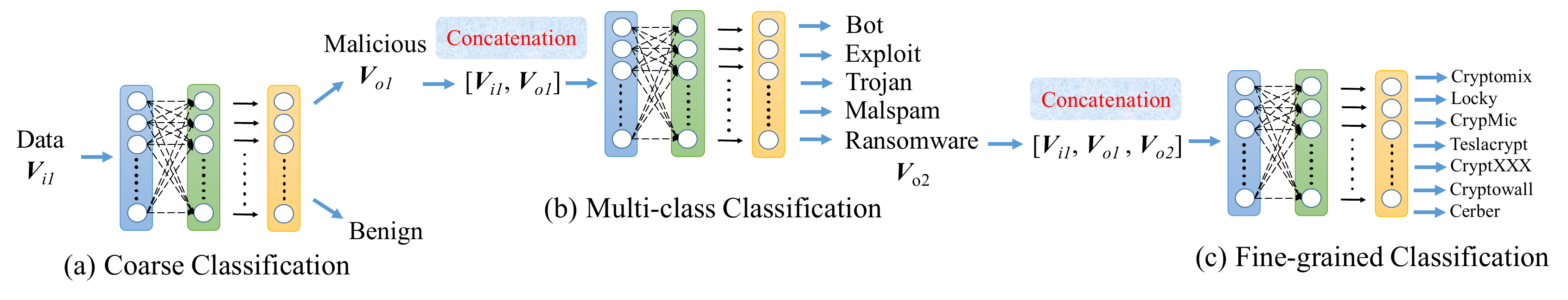}
    \vskip -1em
    \caption{This model performs a 12-class classification in three stages. Firstly, the model performs a coarse classification which preliminarily classifies the data into the class of benign flows and malicious flows. Secondly, the model conducts a multi-class classification on the malicious flows, which classifies them into 5 different categories according to the attack behavior. Lastly, the model executes a fine-grained classification on Ransomware family.}
  \end{figure*}

\section{Methodology}

\vspace {-1mm}

\subsection{Quantity Dependent Backpropagation (QDBP)}
Backpropagation is a gradient-based method to train neural networks. The mathematical formula is given by:

\vspace {-1.5mm}

  \begin{equation}
    \theta_i^{l^+} = \theta_i^l - \eta\times\frac{\partial Loss}{\partial \theta_i^l}
  \end{equation}

\vspace {-1.5mm}

where \(\theta_i^l\) and \(\theta_i\textsuperscript{l\textsuperscript{+}}\) represent the $i^{th}$ parameter in layer \(l\) and the updated parameter respectively, $\eta$ is the learning rate, $\frac{\partial Loss}{\partial \theta_i^l}$ refers to the partial derivative of \(Loss\) with respect to \(\theta_i^l\).

However, neither backpropagation nor adaptive learning rate approaches can't reflect the sensitivity toward minority classes whether applying online learning or batch learning mechanism since both methods treat each class evenly and are only sensitive to gradient. When imbalanced data distribution occurs, these methods will result in the inability of the model to learn from the minority classes due to gradient dilution issue.

To mitigate this issue, we introduce a vector $\boldsymbol F$ into backpropagation (equation (3)) and propose a QDBP algorithm which takes the disparity between classes into consideration and shows different sensitivities toward different classes. The mathematical formula is given by:

\vspace {-3mm}

\begin{equation}
    \theta_i^{l^+} = \theta_i^l - \eta\cdot\boldsymbol F\cdot\nabla Loss
\end{equation}

\vspace {-3mm}

\begin{equation}
    \boldsymbol F = \bigg [ \frac{c_1}{n_1},  \frac{c_2}{n_2}, ...,  \frac{c_N}{n_N} \bigg ]
\end{equation}

\vspace {-4mm}

\begin{equation}
    \nabla Loss = \bigg [ \frac{\partial Loss_1}{\partial \theta_i}, \frac{\partial Loss_2}{\partial \theta_i}, ..., \frac{\partial Loss_N}{\partial \theta_i} \bigg ]^T
\end{equation}

where $\boldsymbol F$ is a row vector, \(N\) represents the cardinality of the training data set, $n_i$ represents the cardinality of the class where the $i^{th}$ data belongs to, and $c_i$ is the pre-selected coefficient for the class where the $i^{th}$ data belongs to. The adjustment of \(c_i\) depends on the sensitivity of the model to the gradients contributed by the data belonging to class \(M_i\). $\nabla Loss$ is a column vector composed of the partial derivatives of loss with respect to $\theta$ contributed by each data. For example, if the $i^{th}$ data belongs to class \(M_k\), and the pre-selected coefficient for class \(M_k\) is c, then \(n_i\) = \(|M_k|\) and \(c_i\) = c.

From equation (4), we can show that:

\vspace {-4mm}

\begin{equation}
    \theta_i^{l^+} = \theta_i^l - \eta\sum_{M_t\in M}^{}\frac{c_{M_t}}{n_{M_t}}\sum_{A_{tk} \in M_t}^{}\frac{\partial Loss_{A_{tk}}}{\partial \theta_i}
\end{equation}

\begin{equation}
    M = \bigcup_{t=1}^{|M|} M_t \hskip 0.5em , \hskip 1.5em  M_t = \bigcup_{k=1}^{|M_t|} A_{tk}
\end{equation}

where \(M\) represents the training data set, \(M_t\) refers to the $t^{th}$ class in \(M\), and \(A_{tk}\) is the $k^{th}$ data of \(M_t\).

By selecting $c_{M_t}$ to 1, $\forall M_t \in M$,\\
$\frac{1}{n_{M_t}}\sum_{A_{tk} \in M_t}^{}\frac{\partial Loss_{A_{tk}}}{\partial \theta_i}$ can be viewed as the normalized equivalent gradient computed by each data $A_{tk},$ $\forall A_{tk} \in M_t$ since $n_{M_t} = |M_t|$.

Under this mechanism, gradient dilution issue will be mitigated since the total gradient contributed by each class accounts for the same proportion in terms of contribution. In our experiment, to reflect the sensitivity of the minority class, the pre-selected coefficient for the minority class is set to 1.2 while others remain 1.

\subsection{Tree-Shaped Deep Neural Network (TSDNN)}
To mitigate the imbalanced data issue, we propose an end-to-end trainable TSDNN model (Fig. 1) which classifies the data layer by layer. We define each model in TSDNN as a nodal network and the links between nodal networks are called bridges. We adopt cross entropy as the loss function and apply QDBP to each nodal network to optimize the performance.

Unlike the architecture of a multi-layer perceptron model which is a static structure, the model we proposed grows and expands dynamically as the classification goes more detailed.

As shown in Fig. 1 (a), we firstly combine all the malware data into a group labeled "Malicious" and perform a coarse classification separating the malicious flows from the benign ones. Under this scenario, the number of the two classes does not exist a large disparity compared to the distribution of the entire data set and therefore the gradient dilution issue will hardly alter the classification. After the preliminary classification, the data labeled "Malicious" will firstly be transferred into a concatenation stage then fed into a subsequent nodal network, which is the child of the current nodal network, to further perform a detailed classification.

In the second layer of TSDNN (Fig. 1 (b)), we would like to classify the malicious flows according to their attack behavior. As far as the authors' understanding, the attack behavior of the malware can be categorized into 5 categories as listed at the output end in Fig. 1 (b). Thus, the output dimension of the second nodal network is set to 5. Instead of directly feeding the output vector $\boldsymbol V_{o1}$ of the malicious flow calculated in the first layer into current nodal network, the output vector $\boldsymbol V_{o1}$ will firstly undergo a concatenation process where we concatenate the input vector $\boldsymbol V_{i1}$ of the malicious flow in the first layer with the output vector $\boldsymbol V_{o1}$ computed by the first nodal network and regard the resulting vector $\boldsymbol V_{i2} = [\boldsymbol V_{i1}, \boldsymbol V_{o1}]$ as the input vector of the second nodal network. Namely, the input vector $\boldsymbol V_{i2}$ of the second nodal network will consist of two parts, the original vector $\boldsymbol V_{i1}$ of the malicious flow and the learned feature $\boldsymbol V_{o1}$ computed by the first nodal network. For example, if the input vector $\boldsymbol V_{i1}$ of the first nodal network is $\boldsymbol V_{i1} = [u_{i1}, u_{i2}, u_{i3}]$ and the corresponding output vector $\boldsymbol V_{o1}$ is $\boldsymbol V_{o1} = [u_{o1}, u_{o2}]$, the input vector of the second nodal network will be given by $\boldsymbol V_{i2} = [u_{i1}, u_{i2}, u_{i3}, u_{o1}, u_{o2}]$. However, as the classification goes finer, the imbalanced data distribution will become more conspicuous meaning that the gradient dilution issue will influence the performance. By suitably selecting the coefficients \(c_i\) for each class, QDBP can still improve the performance. Afterwards, the data labeled "Ransomware" will be fed into a succeeding nodal network which is the child of the current nodal network to further conduct a fine-grained classification.

Lastly, we conduct a fine-grained classification demonstrated in Fig. 1 (c). In this layer, we classify the data of the Ramsomware family in particular. As described, the input vector $\boldsymbol V_{i3}$ of this nodal network will incorporate the knowledge learned from the previous nodal networks. In other words, we concatenate the output vectors $\boldsymbol V_{o1}$ and $\boldsymbol V_{o2}$ computed by the first and second nodal networks respectively with the input vector $\boldsymbol V_{i1}$ of the first nodal network and consider the resulting vector $\boldsymbol V_{i3} = [\boldsymbol V_{i1}, \boldsymbol V_{o1}, \boldsymbol V_{o2}]$ as the input vector of the current nodal network. Likewise, imbalanced data distribution arises in this layer. By selecting suitable coefficients \(c_i\) for each class, we can still conquer the difficulty caused by the gradient dilution issue.

Apart from the multi-layer perceptron which classifies data altogether, we classify the data in a layer-wise manner which can diminish the disparity between classes when analyzing the data distribution in each layer. Furthermore, the input vectors $\boldsymbol V_{i2}$ and $\boldsymbol V_{i3}$ of the second and the third nodal networks are not directly fed by the output vectors $\boldsymbol V_{o1}$ and $\boldsymbol V_{o2}$ of the preceding nodal networks. Instead, the advantage of such a hierarchical classification is that we can generate new features $\boldsymbol V_{o1}$ and $\boldsymbol V_{o2}$ in each layer and incorporate the learned features into the next level's input vector to further provide more information. Moreover, since TSDNN is end-to-end trainable meaning that all the nodal networks are trained simultaneously, TSDNN can tune the output vector of each nodal network through QDBP to improve the erroneous classification in the learning phase.  Another advantage is that TSDNN can grow and expand dynamically, which differs from other neural networks structurally. Namely, the more detailed the classification is desired, the more layers and nodal networks will be involved in TSDNN.

\section{Experiments}

\subsection{Data Collection}
The malware traffic used in the experiments was collected from September 2016 to May 2017 so as the benign traffic. We record the first 6-minute network behavior of the malicious samples obtained from VirusTotal(https://www.virustotal.com) under sandbox.

The malicious data we collected are categorized into 5 different classes where each class represents different attack behaviors. We further label the data of the Ransomware family:Cryptomix, Locky, CrypMic, Telslacrypt, CryptXXX, Cryptowall, Cerber, according to the official name listed in VirusTotal. Therefore, our data set is now composed of 12 different classes as illustrated in Table \RN{1}.


\begin{flushleft}
  \begin{table}[ht]
    \centering
    \vskip -2em
    \caption{Data Statistics}
    \begin{tabular}{l | r | r}
     Class & Number of Flows & Size\\ [1ex] 
     \hline\hline
     Benign & 246,015 & 560.2 MB\\ 
     Bot & 99 &  6.5 MB\\ 
     Exploit & 349 & 32.5 MB\\ 
     Trojan & 3,085 & 18.1 MB\\ 
     Malspam & 3,612 & 142.1 MB\\ 
     Cryptomix & 90 & 2.0 MB\\
     Locky & 229 & 9.3 MB\\
     CrypMic & 390 & 14.3 MB\\ 
     Telslacrypt & 755 & 26.5 MB\\
     CryptXXX & 1,259 & 44.7 MB\\ 
     Cryptowall & 2,864 & 34.7 MB\\ 
     Cerber & 23,260 & 23.5 MB\\  [1ex] 
     \hline\hline
     Total & 282,007 & 914.4 MB 
    \end{tabular}
    \label{table:1}
  \end{table}
\end{flushleft}

\vspace {-20mm}

\begin{flushleft}
  \begin{table*}[!t]
  \centering
  \caption{Accuracy and precision of different approaches}
  \begin{tabular}{l | r | r} 
    Method & Accuracy & Precision\\ [1ex] 
    \hline\hline
    DNN + Backpropagation & 59.08\% & 8.33\%\\
    DNN + Oversampling (10000 samples/class) \cite{pears2014synthetic}  & 85.18\%  & 65.9\%\\
    DNN + Undersampling (45 samples/class) \cite{liu2009exploratory} & 68.89\% & 49.45\%\\
    DNN + Incremental Learning \cite{kulkarni2014incremental} & 78.84\% & 71.23\%\\
    DNN + QDBP & 84.56\%  & 62.3\%\\
    SVM (RBF) & 83.87\% & 38.8\%\\
    Random Forest & 98.9\% & 68.25\%\\
    \textbf{TSDNN + QDBP} & \textbf{99.63\%} & \textbf{85.4\%}\\ [1ex] 
    \end{tabular}
    \label{table:4}
  \end{table*}
\end{flushleft}

\subsection{Feature Extraction}

\subsubsection{Connection Records}
Referring to Williams et al. \cite{williams2006preliminary}, Internet protocols and connection statistics are significant to the identification of traffic and flow. Thus, we extract the connection records like port, IP, and protocol information in traffic flows from the transport layer. In addition, according to Anderson et al. \cite{anderson2016deciphering} and Tseng et al. \cite{ransomwareDL}, the information of the HTTP requests and TLS handshake will play a crucial role in traffic identification. These information can be obtained in TCP payloads.

\subsubsection{Network Packet Payload}
TCP and UDP payloads always contain the transmitted data, and the former are sometimes sent encrypted under TLS or SSL protocols. In order to trace the malicious flows and the accompanied malware, the transmitted data of all payloads in a flow is important and should be taken into consideration. Hence, we extract all the payloads from each packet where the length of each packet ranges from 0 to 1,500 bytes.

\subsubsection{Flow Behavior}
Flow behavior is described as a way to monitor the sending process of packets whereas each malware family possesses different flow behaviors. According to Moore et al. \cite{moore2013discriminators}, they proposed nearly 250 discriminators to classify the flow record. Among these discriminators, inter-arrival time of each packet in a flow plays the most important role in terms of classification. Besides, from McGrew et al. \cite{mcgrew2016enhanced}, they used Markov matrix to store the relationship between sequential packets. Inspired by their methods, we apply Markov transition matrix to represent the flow behavior and document the before-and-after relationship.

\subsection{Malicious Flow Detection and Layer-wise Analysis}

\subsubsection{Analysis on the Malicious Flow Detection}
Firstly, we would like to test if our model is able to distinguish malicious flows from the benign ones. Referring to Tseng et al. \cite{ransomwareDL}, their result could achieve a 93\% accuracy in malicious flow detection using only HTTP headers and TCP payloads. However, after including the new features mentioned in the previous section, we can achieve an accuracy of 99\% under vanilla backpropagation. If we further apply QDBP, the recognition rate can even improve to 99.7\%. This evidence shows the improvement from the past work \cite{ransomwareDL} to current and justifies that it is the feature that results in the improvement under same training mechanism. On the other hand, the subtle improvement contributed by QDBP indicates that when the number of data in the minority class is large to some extent, 35,992 in our task, the imbalanced data distribution won't be significant since the quantity is large enough to construct a robust model.

\subsubsection {Comparison of Different Approaches on Flow Classification}
In addition to detecting malicious flows, we would like to further classify the malicious flows more detailed into 11 different classes. In our task, there are 246,015 flows belong to the "Benign" class while "Cryptomix" class only contains 90 flows. Such disparity will result in an insensitive model which will misclassify the data into the majority class if classifying the data altogether through vanilla backpropagation. The result is illustrated in Fig. 2. Fig. 2 is the confusion matrix result of a 12-class classification, confusion matrix is a concrete way to present the result of a multi-class classification where true label represents the the ground true label while the predicted label is the result that given by the model. For example, if a data labeled "Bot" but predicted "Exploit", then the block corresponding to that classification will be increased by 1. An ideal confusion matrix will have a dark color in the diagonal while others bright. From the confusion matrix shown in Fig. 2, we can observe that all testing data is classified into the "Benign" class, where the shade of the color of each block indicates the probability of the corresponding classification.

\vspace {-5mm}

\begin{figure}[ht]
  \centering
  \includegraphics[scale=.3]{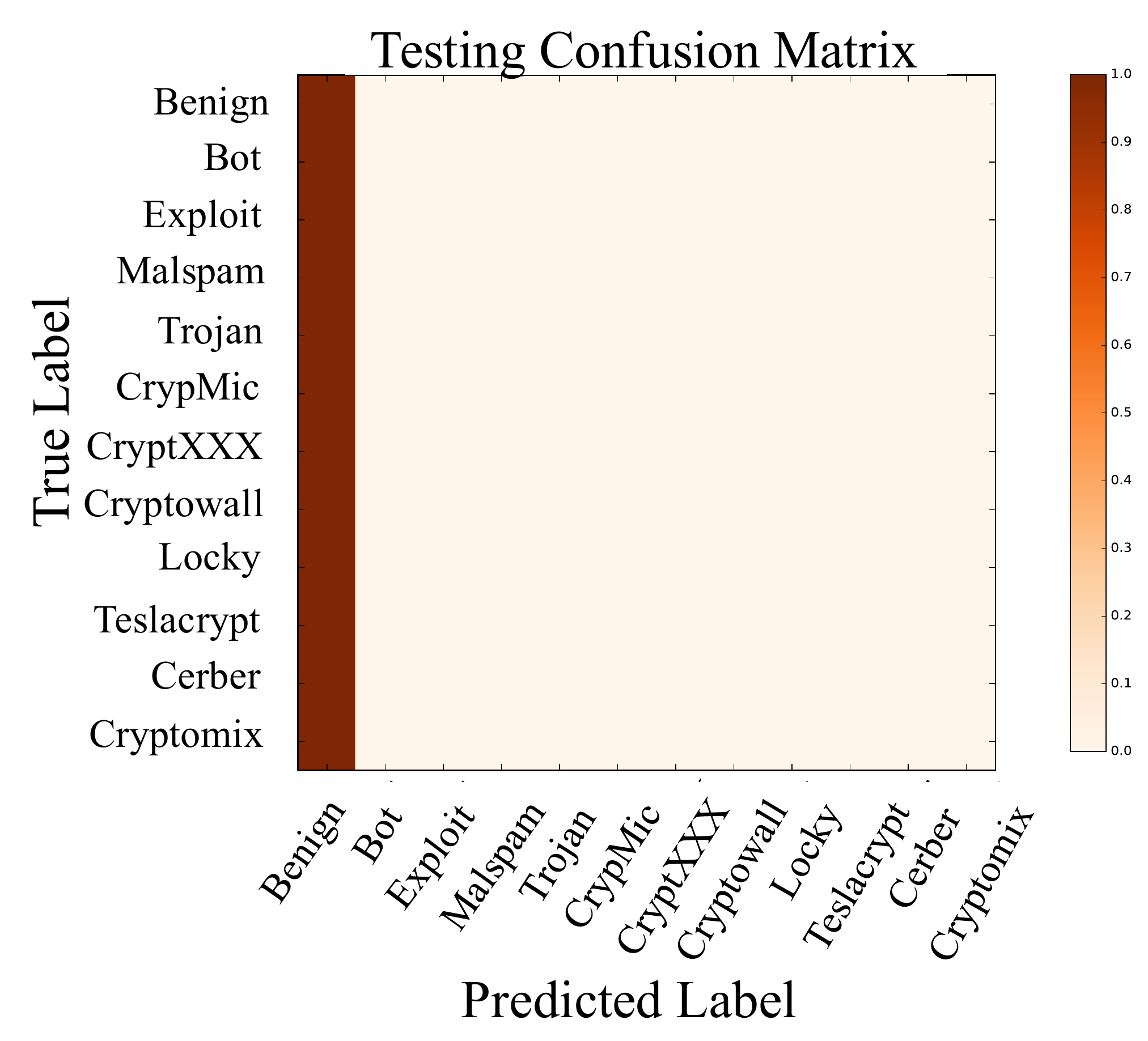}
  \vskip -1em
  \caption{Confusion matrix of a traditional DNN with 3 hidden layers. The DNN is trained with the vanilla backpropagation algorithm.}
\end{figure}

However, from Table \RN{2}, the accuracy of vanilla backpropagation can still achieve 59\% even if the model bias toward the majority class since the testing data set contains roughly 30 thousand benign samples and 20 thousand malicious samples. This shows that the accuracy is not a comprehensive evaluation index because it can easily be manipulated by adjusting the distribution of the testing data set.

To evaluate the model in a more precise manner, we adopt another performance metrics called average precision and the mathematical formula is given by:

\vspace {-2mm}

\begin{equation}
    Precision_{avg} = \frac{1}{N}\sum_{i=1}^{N}\frac{TP_{i}}{TP_{i}+FP_{i}}
\end{equation}

\vspace {-2mm}

where \(N\) represents the number of classes, $N=2$ in the coarse classification and $N=12$ in the 12-class classification. $TP_i$ and $FP_i$ stand for True Positive and False Positive of class \(i\) respectively. Precision is one of the performance metrics which represents the ratio between the correctly classified samples and the total data in each class. This performance metrics won't be seriously affected by the existence of the majority class since it calculates the average precision of each class. Though the cardinality of the "Benign" class is extremely large, this performance metrics won't bias toward the majority class completely since it treats each class evenly. From Table \RN{2}, the precision of vanilla backpropagation is 8.33\% which can reasonably support the phenomenon shown in Fig. 2.

To mitigate the influence of the imbalanced data distribution, we structure our neural network to be hierarchical and classify the data layer by layer. Under this setting, the data distribution in each layer will be more balanced compared to prior methods. We can reduce the quantity ratio between the majority class and the minority class from $\frac{216,015}{45}$ to $\frac{14,423}{50}$ which nearly shrinks the ratio by 17 times. Besides, we apply QDBP in each layer to further improve the gradient dilution issue. 

\vspace {-5mm}

\begin{figure}[ht]
  \centering
  \includegraphics[scale=.3]{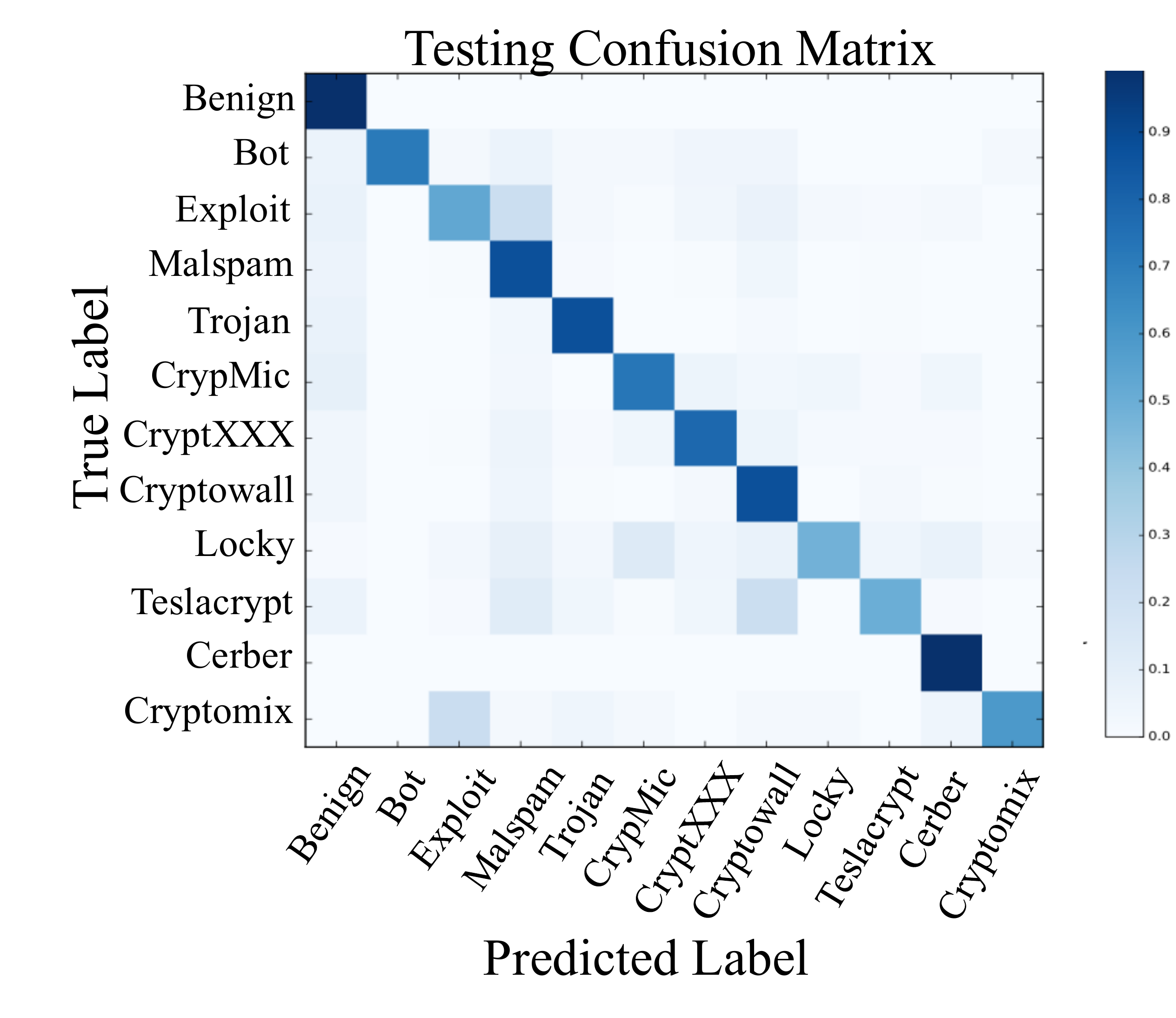}
  \vskip -1em  
  \caption{Confusion matrix of TSDNN. TSDNN is trained with QDBP in each nodal network.}
\end{figure}

From the confusion matrix shown in Fig. 3, our method enhances the performance of classifying the entire imbalanced data set not only in the classification accuracy where the accuracy is improved from 59.08\% to 99.63\% but in the sensitivity of the model where the precision is improved from 8.33\% to 85.4\%. We further conclude that the superior performance in training an imbalanced data set has to ascribe to two factors, the TSDNN model which diminishes the disparity in each layer and the proposed QDBP algorithm which mitigates the gradient dilution issue.

We also compare the result with the state-of-the-art methods \cite{pears2014synthetic}, \cite{liu2009exploratory}, \cite{kulkarni2014incremental} mentioned in the related work section. As illustrated in Table \RN{2}, the performance of the vanilla backpropagation is seriously affected by the gradient dilution issue because the cardinality of "Bot" and "Cryptomix" classes is much smaller than that of "Benign" class. Even though the state-of-the-art methods can improve the performance of classification, these methods still can't resolve the gradient dilution issue. If incorporating QDBP into TSDNN, the recognition rate can even achieve 99.63\% in accuracy and 85.4\% in precision which outperforms other approaches whether in accuracy or in precision.


\subsection{Partial Flow Detection}
For the sake of real-time detection, we devise an experiment by dividing each attack into fractions and only consider a portion of the data to test the potentiality of being a malware.

As illustrated in Fig. 4, the recognition rate of malicious flow detection ascends as the portion of the data increases. Besides, from the result shown in Fig. 4, we can conclude that our model is able to distinguish the malicious flow by only considering the first 5 percent of the entire flow which shows the possibility of a real-time detection since the model can perceive the potential threats in the very beginning of the process without analyzing the entire flow.

\vspace {-2mm}

\begin{figure}[ht]
  \centering
    \includegraphics[scale=.3]{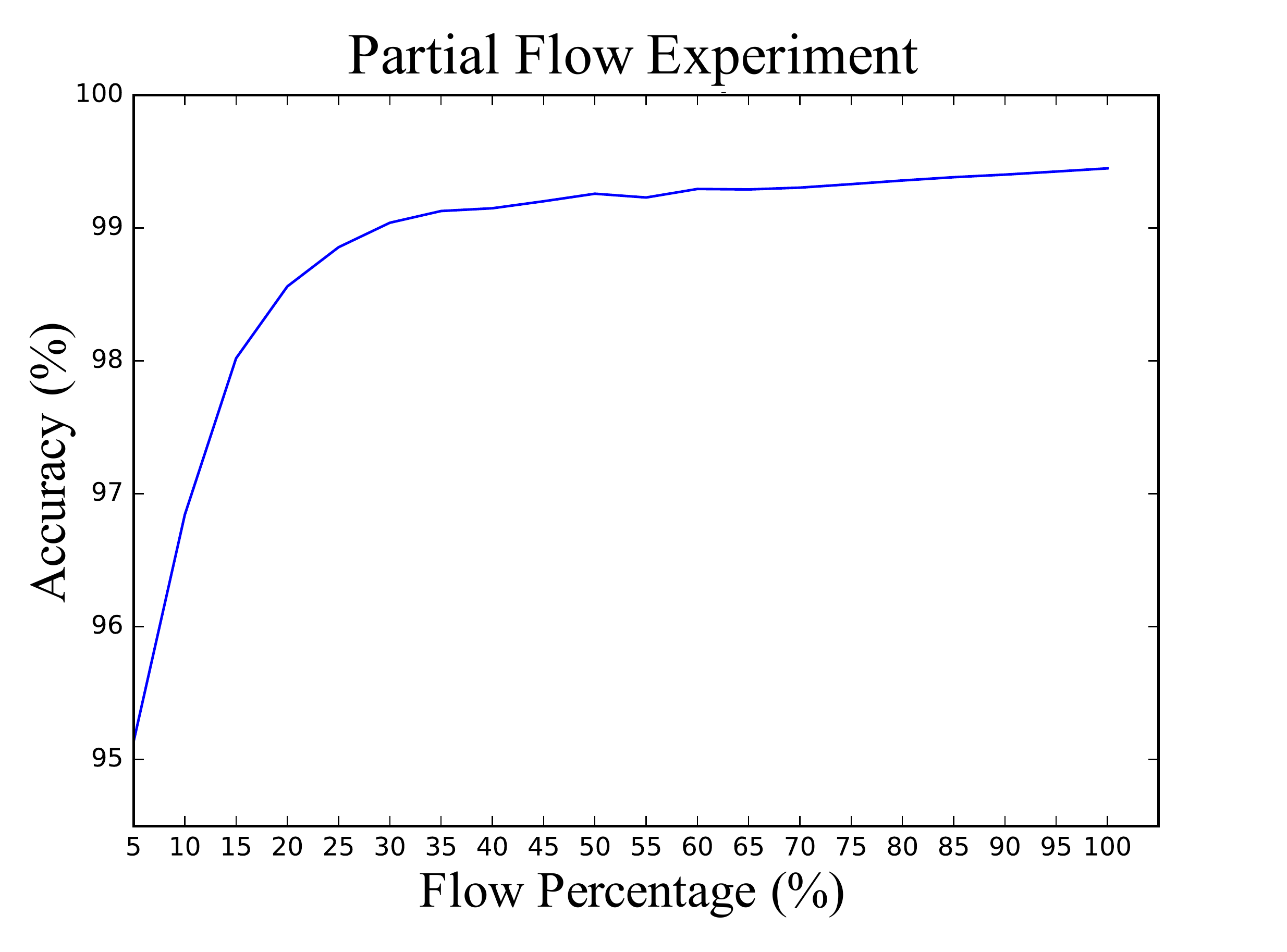}
  \caption{In the 6 minutes of each attack, our model can achieve 95\% accuracy perceiving malicious flows in the first 5 percent of the duration, namely 18 seconds. The first part of flows in the connection gives enough clues for alerting malicious attacks.}
\end{figure}

\subsection{Zero-shot Learning}
There are various kinds of malware existing in the cyber society. However, it is impossible to collect data samples of each family since there are many new variants coming into existence every day. To evaluate the generalization performance of the proposed model, we would like to examine the ability of TSDNN to identify some malware that has never been trained by our model. This kind of scenario is coined as a "zero-shot learning" in machine learning term. Therefore, we collect 14 different kinds of malware (Fig. 5) to evaluate the ability of our model to perceive potential threats and the experimental result is illustrated in Fig. 5.

From the experimental results, we can observe that our model performs reasonably well on recognizing these malware. In real world, each attack usually has several network flows. Once any part of these flows is recognized as a malicious connection, the attack will be blocked and the process will be terminated immediately. This result not only shows the ability to predict potential threats but further justifies that a behavior-oriented approach to detect malware is a better way compared to the traditional signature-based methods, which also accounts for the generalization capability of deep learning.

\vspace {-2mm}

\begin{figure}[ht]
  \centering
    \includegraphics[scale=.4]{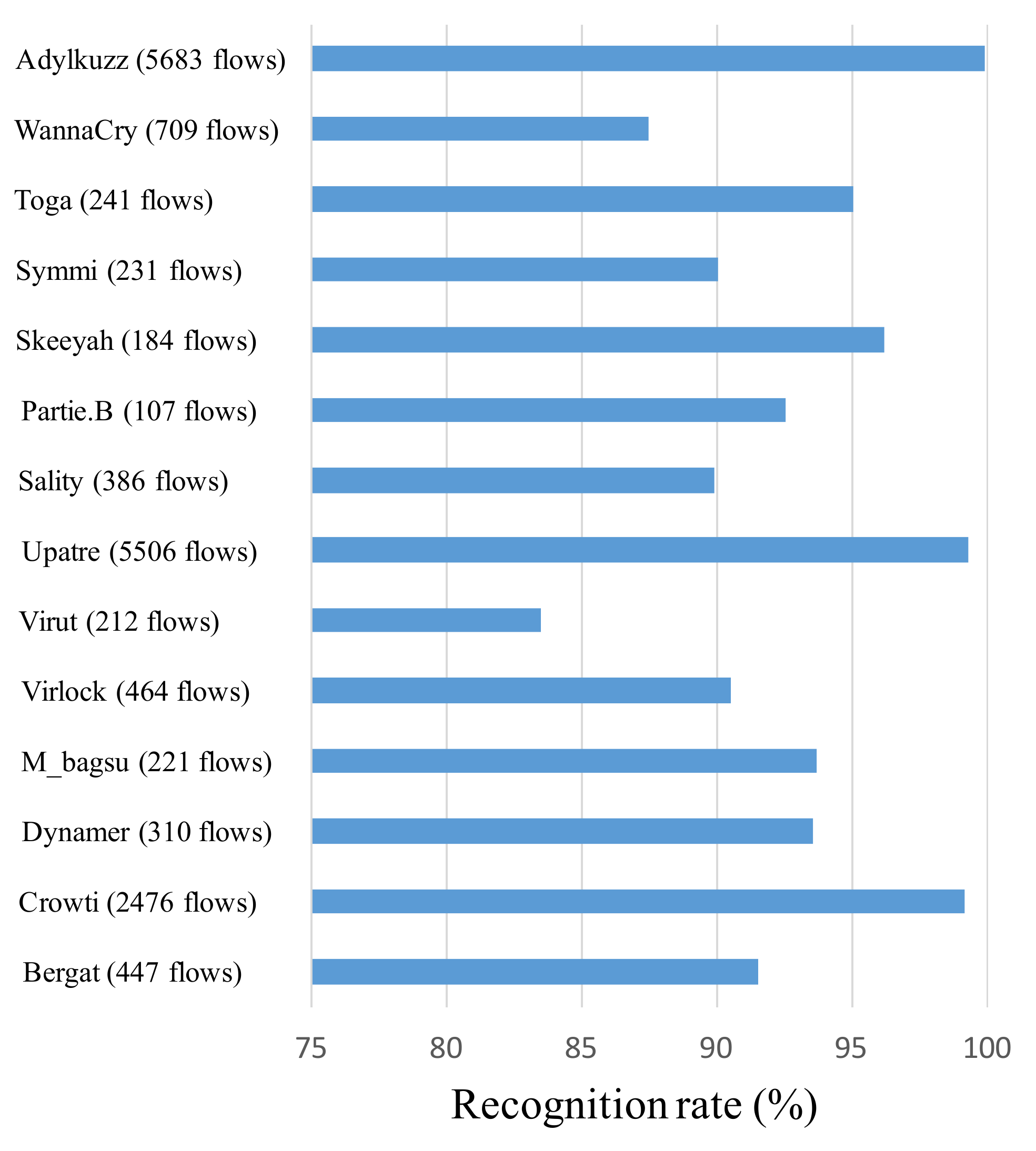}
  \vskip -1em
  \caption{The new malware data set contains 14 different families which are different from that used in training TSDNN.}
\end{figure}

\section{Conclusion}
We propose an end-to-end trainable TSDNN model along with a QDBP algorithm, which enables the model to memorize and learn from the minority class, to perform malicious flow detection. Unlike the multi-layer perceptron architecture which classifies the data altogether, we classify the data in a layer-wise manner. We further conduct a partial flow detection which is meant to seek for the possibility of real-time detection by only considering a portion rather than the entire flow. From the experimental result, real-time detection can truly be fulfilled. To evaluate our model's ability to detect potential malware, we execute an experiment on testing the malware that has never been trained by our model. The experimental results show that our model is able to accurately detect the potential malware at a superior performance which also justifies that behavior-oriented approach is better than signature-oriented methods when detecting malware.

\section{Acknowledgement}
This work is supported by the Ministry of Science and Technology, Taiwan, under Grant MOST 106-3114-E-002-005 and MOST 106-2627-M-002-023.

\bibliography{main}

\end{document}